\documentclass[letterpaper]{article} 
\usepackage{aaai2027}  
\usepackage[hyphens]{url}  
\usepackage{graphicx} 
\usepackage{natbib}  
\usepackage{caption} 
\usepackage{algorithm}
\usepackage{algorithmic}
\usepackage[table]{xcolor}
\usepackage{newfloat}
\usepackage{listings}
\DeclareCaptionStyle{ruled}{labelfont=normalfont,labelsep=colon,strut=off} 
\floatstyle{ruled}
\newfloat{listing}{tb}{lst}{}
\floatname{listing}{Listing}

\usepackage{booktabs}
\usepackage{amsmath}
\usepackage{multirow}
\nocopyright 

\title{Training-Free VLM Personalization via Calibrated Residual Decoding}
\author {
    Jiaao Yu\textsuperscript{\rm 1},
    Yujian Ma\textsuperscript{\rm 1}, 
    Xianming Hu\textsuperscript{\rm 1}, 
    Pengran Wang\textsuperscript{\rm 1}, 
    Ang Li\textsuperscript{\rm 2}, 
}
\affiliations {
    \textsuperscript{\rm 1}East China Normal University, Shanghai, China\\
    \textsuperscript{\rm 2}Chang'an University, Shanxi, China\\
}

\begin{document}

\maketitle

\begin{abstract}
Vision-language models can be personalized in a training-free manner by directly providing user profiles, preferences, or visual references at inference time, without updating model parameters. However, direct personalized prompting does not guarantee that the model will reliably exploit such evidence. The predictive distribution under the positive user profile often mixes two sources: personalized signals genuinely supported by the current profile, and the model's generic visual or linguistic priors. As a result, from the positive-profile response alone, it is difficult to determine whether a high-confidence answer is supported by the user profile or merely reflects the model's default preference. To address this problem, we propose a training-free calibrated residual decoding framework. Given the same image and question, we construct three evidence conditions: a positive profile $E^{+}$, a counterfactual profile $E^{-}$, and an empty profile $E^{0}$. Our method keeps the prediction under $E^{+}$ as the anchored base, and explicitly estimates the marginal contribution of personalization from score differences across the three conditions. We further introduce normalized-entropy-based uncertainty calibration, allowing the strength of personalized enhancement to adapt to the reliability of the residual signal. Experiments on MMPB, YoLLaVA, and MyVLM show that the proposed method improves personalized multimodal understanding without fine-tuning, with consistent gains on identity-sensitive visual personalization tasks. Additional analysis shows that entropy calibration stabilizes residual decoding when the contrastive personalization signal is uncertain.
\end{abstract}


\section{Introduction}

Vision-language models (VLMs) have achieved remarkable progress in image understanding, visual question answering, and multimodal reasoning, enabling models to generate plausible responses conditioned on visual content and natural language instructions. However, real-world human-AI interaction is rarely fully generic: different users often have different preferences, background knowledge, long-term habits, and visual experiences. For the same image and the same question, the answer that matters to one user may depend on their personal preferences, identity cues, or existing profile. In scenarios such as personalized recommendation, companion assistants, adaptive shopping, personal object recognition, and long-term multimodal interaction, a model must not only understand what is present in the image, but also determine what is relevant to a particular user. Personalized multimodal understanding is therefore becoming an important capability for VLMs to evolve toward human-centered intelligent systems.

\begin{figure}[t]
\centering
\includegraphics[width=0.46\textwidth]{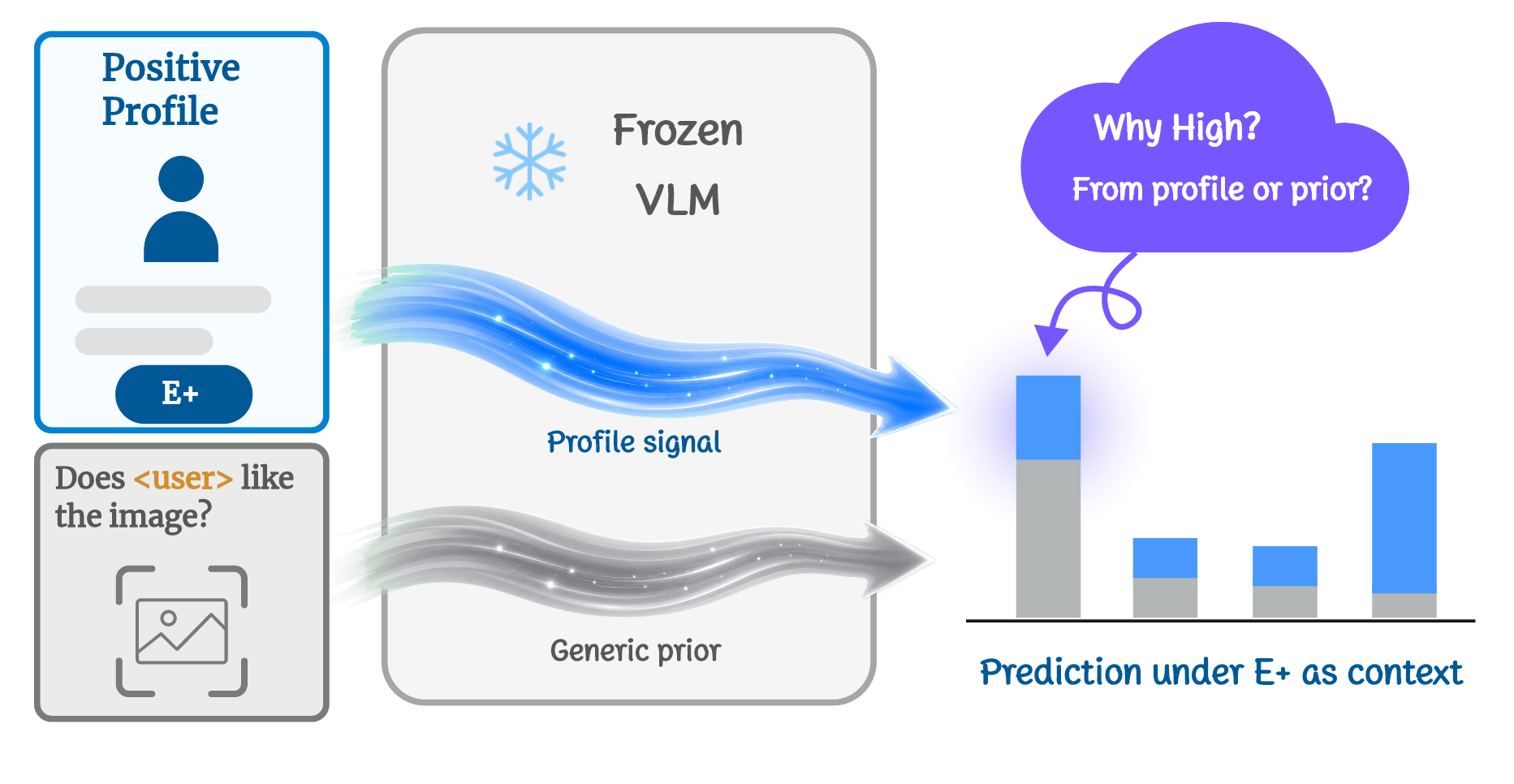} 
\caption{Ambiguity of positive-profile prompting. The output distribution under the positive profile $E^{+}$ mixes profile-supported signals with the VLM's generic prior. From $E^{+}$ alone, a high-scoring response may be difficult to attribute to either source.}
\label{fig1}
\end{figure}

Recent studies have begun to explore personalized VLMs \cite{zhu2025internvl3,wang2024qwen2,bai2025qwen3,liu2023visual,dai2023instructblip}. Existing approaches typically rely on additional training, user memory modules, or curated personalized data to adapt models to user-specific preferences and concepts. For example, MyVLM \cite{alaluf2024myvlm}, Yo'LLaVA \cite{nguyen2024yo}, and PersonaVLM \cite{nie2026personavlm} inject personalization by learning user-specific concept representations, personalized latent tokens, or long-term memory alignment. Although these methods improve adaptation to specific users, they also introduce training cost, privacy concerns, and deployment complexity. In contrast, a lightweight alternative is to directly provide user profiles, historical preferences, or reference images as context to an existing multimodal model. Such prompting-based personalization does not require parameter updates and is therefore easier to deploy on top of frozen VLMs.

However, directly supplying personalized context does not guarantee that the model will use it stably or appropriately. As shown in Figure ~\ref{fig1}, under the positive-profile condition, the resulting predictive distribution mixes two sources: personalized signals genuinely supported by the current user profile, and the model's generic visual or linguistic priors. From the positive-profile response alone, we cannot determine whether a high-scoring answer is preferred because it is truly supported by the current user profile, or simply because the model tends to favor it regardless of personalization. As a result, absolute scores, confidence, or margins under the positive profile alone are insufficient to decide when personalized enhancement should be trusted. To reliably exploit personalization in a training-free setting, we need additional reference conditions that explicitly measure the marginal effect of user-specific evidence.

This observation makes our setting related to, yet fundamentally different from, prior contrastive decoding methods such as Context-Aware Decoding (CAD) \cite{shi2024trusting}, Visual Contrastive Decoding (VCD) \cite{leng2024mitigating}, and DoLa \cite{chuang2024dola}. These methods demonstrate that comparing outputs across different conditions can improve model behavior by suppressing hallucinations, reducing visual bias, or exposing shallow priors. However, they typically treat the removed condition as a noise source or bias source, with the goal of weakening unreliable evidence. Personalization presents a different scenario: the positive user profile is not a bias to be removed, but legitimate evidence that should be preserved. Our goal is therefore not to suppress personalization, but to calibrate how its effect should be incorporated into the final prediction while retaining the positive-profile response as the base. In other words, what we need is not de-personalization, but calibrated personalized enhancement.

Motivated by this perspective, we revisit personalized multimodal decoding and propose a training-free calibrated residual decoding framework. Given the same image and question, we construct three evidence conditions: a positive user profile $E^{+}$, a counterfactual user profile $E^{-}$, and an empty profile $E^{0}$. Here, $E^{+}$ represents the current user's true preferences or identity cues, $E^{-}$ represents the preferences or identity cues of an alternative user, and $E^{0}$ denotes a generic condition without personalized evidence. By comparing the scores assigned to the same candidate response across these three evidence conditions, we construct a personalized residual signal while preserving the prediction under the positive profile as the anchored base. Furthermore, we introduce an uncertainty calibration mechanism based on normalized entropy, allowing the model to adaptively modulate the strength of personalization according to the reliability of the residual signal. This yields a calibrated response score that can be directly used for personalized decoding in VLMs.


The main contributions of this work are as follows:
\begin{itemize}
    \item We propose a training-free personalized multimodal decoding framework that anchors on the positive user profile and explicitly estimates personalized residual signals from counterfactual and empty profiles.
    \item We introduce an entropy-based uncertainty calibration mechanism that adaptively controls the strength of personalized enhancement according to the reliability of the residual signal.
    \item We validate the proposed framework on multiple personalized multimodal benchmarks, showing that it improves personalized understanding without model fine-tuning and provides a unified inference-time view across text-profile and image-reference personalization.
\end{itemize}


\begin{figure*}[t]
\centering
\includegraphics[width=0.97\textwidth]{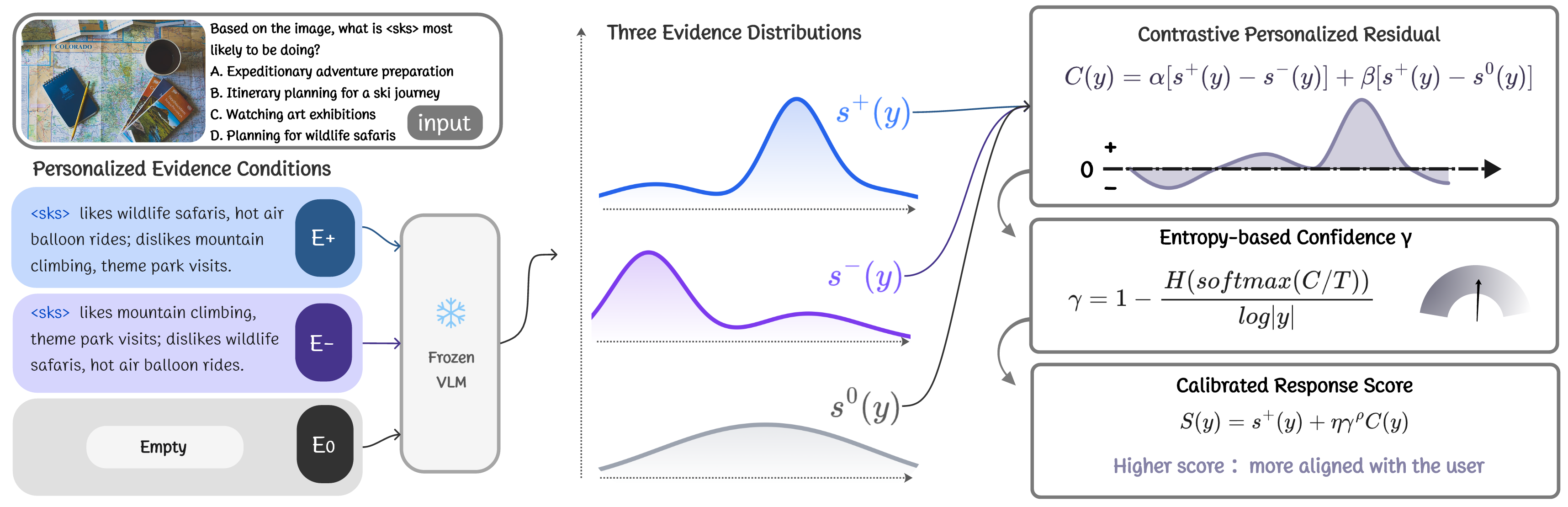} 
\caption{Our decoding framework estimates the marginal contribution of personalization via three evidence conditions ($E^+$,$E^-$,$E^0$), with entropy-based calibration controlling the residual strength.}
\label{fig2}
\end{figure*}

\section{Related Work}

\paragraph{Personalized vision-language models.}
Personalization has been widely explored in generative and language models \cite{kumari2023multi,ning2025user,zhang2024personalized,guo2023animatediff,ding2023diffusionrig}.
Recent work has explored how to equip VLMs and MLLMs with user-specific knowledge. 
MyVLM~\cite{alaluf2024myvlm} and Yo'LLaVA~\cite{nguyen2024yo} learn personalized concept representations or latent visual tokens from user-provided examples, while PVIT~\cite{pi2025personalized} improves personalization through visual instruction tuning. 
To reduce the need for model adaptation, PLVM~\cite{pham2025plvm} studies tuning-free referential concept alignment, and R2P~\cite{das2025training} retrieves user-specific fingerprints for personalized concept reasoning. 
RAP~\cite{hao2025rap} further introduces retrieval-augmented personalization for MLLMs, and PersonaVLM~\cite{nie2026personavlm} builds long-term multimodal memories for persistent user profiles. 
Among existing benchmarks, MMPB~\cite{kim2026mmpb} evaluates profile-dependent preference and identity reasoning, while MyVLM and Yo'LLaVA focus on personalized visual recognition from user-provided reference images.

Our work is complementary to these studies: while prior methods mainly address how personalized evidence is learned, retrieved, or maintained, we study how a frozen VLM should use already-provided evidence during decoding.


\paragraph{Inference-time Decoding.}
Inference-time adaptation broadly includes prompting~\cite{wei2022chain,song2023llm}, retrieval~\cite{li2023contrastive}, and decoding-time control~\cite{chen2023accelerating}.
Inference-time methods have been widely used to adapt frozen language and multimodal models without parameter updates, including in-context learning, retrieval augmentation, decoding-time control, and contrastive decoding.
Our work is also related to contrastive decoding, which improves generation by comparing output distributions from different models, contexts, inputs, or internal states. 
Early methods such as Contrastive Decoding~\cite{li2023contrastive}, DExperts~\cite{liu2021dexperts}, and related contrastive generation methods~\cite{su2022contrastive, o2023contrastive} control generation by contrasting expert, weaker, or alternative distributions, while DoLa~\cite{chuang2024dola} contrasts predictions from different model layers. 
Closer to our setting, Context-Aware Decoding~\cite{shi2024trusting} and adaptive variants~\cite{wang2025adacad, kim2024adaptive} compare predictions with and without external context to reduce reliance on parametric priors. 
In LVLMs, Visual Contrastive Decoding~\cite{leng2024mitigating} and hallucination-aware decoding methods~\cite{huang2024opera, chen2024halc} mitigate hallucinations through input-level contrast or decoding-time calibration. 
Unlike these methods, which mainly use contrastive signals to suppress undesirable sources, personalized decoding must preserve the positive user profile as legitimate evidence. 
We therefore anchor decoding on the positive-profile prediction and use counterfactual and empty-profile conditions as references for estimating and calibrating the marginal contribution of personalized evidence.


\section{Method}

\subsection{Motivating Observation}

Before introducing our decoding rule, we first examine a limitation of direct positive-profile prompting. A prediction made under the positive profile $E^{+}$ is not necessarily purely driven by personalized evidence; it may still follow the model's non-personalized tendency observed under the empty profile $E^{0}$. To quantify this effect, we analyze two profile-dependent MMPB sub-tasks, preference inconsistency and recognition awareness. We define a \emph{prior-aligned error} as an example where positive prompting is wrong and agrees with the empty-profile prediction, i.e., $\hat{y}^{+}=\hat{y}^{0}\neq y^{*}$.

As shown in Table~\ref{tab:prior_aligned_errors}, positive prompting makes errors on 24.09\% of the combined subset. Among these errors, 84.26\% are prior-aligned, suggesting that many positive-profile failures remain consistent with the model's non-personalized preference. This motivates our use of reference evidence conditions: by comparing the positive profile $E^{+}$ with the empty profile $E^{0}$ and a counterfactual profile $E^{-}$, we can estimate the residual contribution of personalized evidence and use it to calibrate decoding.

\begin{table}[t]
\centering
\small
\setlength{\tabcolsep}{5pt}
\begin{tabular}{lccc}
\toprule
Subset & \#Samples & $E^{+}$ Err. & Prior-Aligned \\
\midrule
Pref. Incon. & 1250 & 25.92 & 78.09 \\
Rec. Aware & 1598 & 22.65 & 89.78 \\
\midrule
Combined & 2848 & 24.09 & 84.26 \\
\bottomrule
\end{tabular}
\caption{
Prior-aligned errors under positive-profile prompting on MMPB. All values except \#Samples are percentages. Prior-Aligned denotes the share of $E^{+}$ errors where $\hat{y}^{+}=\hat{y}^{0}\neq y^{*}$.
}
\label{tab:prior_aligned_errors}
\end{table}

\subsection{Problem Formulation}

We study training-free personalization for vision-language models. Given an image $x$, a question $q$, and a candidate response space $\mathcal{Y}$, the goal is to select the response $y \in \mathcal{Y}$ that best matches both the visual input and the user's personalized evidence. The candidate space can correspond to answer options in multiple-choice QA, or token continuations in open-ended decoding.

In prompting-based personalization, the model is typically conditioned on a positive user profile $E^{+}$, which may contain textual preferences, identity cues, or visual references. A straightforward baseline scores each candidate response only under this positive profile. However, as discussed above, the positive-profile score alone mixes profile-supported signals with the model's generic visual and linguistic priors. We therefore introduce two additional evidence conditions: a counterfactual profile $E^{-}$ and an empty profile $E^{0}$. For each candidate response $y$, we compute:


\begin{equation}
\begin{aligned}
s^{+}(y) &= \log p_{\theta}(y \mid x, q, E^{+}), \\
s^{-}(y) &= \log p_{\theta}(y \mid x, q, E^{-}), \\
s^{0}(y) &= \log p_{\theta}(y \mid x, q, E^{0}),
\end{aligned}
\label{eq:evidence_scores}
\end{equation}
where $p_{\theta}$ denotes the frozen VLM.

\subsection{Personalized Residual from Reference Conditions}

The reference conditions make the marginal effect of personalization observable. The counterfactual profile $E^{-}$ asks whether the same response would also be supported under an alternative user's evidence. The empty profile $E^{0}$ asks whether the same response would remain likely without any personalized evidence. If a candidate receives high support under $E^{+}$ but not under $E^{-}$ or $E^{0}$, the difference indicates a profile-specific signal.

We define a personalized residual:
\begin{equation}
C(y) = \alpha \bigl(s^{+}(y) - s^{-}(y)\bigr) + \beta \bigl(s^{+}(y) - s^{0}(y)\bigr),
\label{eq:residual}
\end{equation}
where $\alpha$ and $\beta$ control the relative strength of the counterfactual and empty-profile comparisons. The first term captures how much the target profile supports $y$ relative to an alternative profile, while the second term captures how much the target profile supports $y$ beyond the model's non-personalized prior.

This residual is not used to replace the positive-profile prediction. Instead, it estimates how the positive-profile score should be adjusted after comparing it with controlled reference conditions, keeping the decoding rule positive-anchored: $E^{+}$ remains the base evidence, while $E^{-}$ and $E^{0}$ provide contrastive measurements of profile-specific support.


\subsection{Entropy-Calibrated Residual Decoding}

The residual signal can vary in reliability across instances. On some examples, the residual is sharply concentrated on one candidate, indicating a clear personalized preference. On others, it is diffuse, suggesting that the reference conditions do not provide a stable direction for personalization. Applying the same residual strength in both cases may over-amplify noisy contrasts.

To calibrate the residual, we convert residual scores into a distribution over candidates:
\begin{equation}
p_{C}(y) = \frac{\exp(C(y)/T)}{\sum_{y' \in \mathcal{Y}} \exp(C(y')/T)},
\label{eq:residual_distribution}
\end{equation}
where $T$ is a temperature parameter. We then compute a normalized confidence coefficient:
\begin{equation}
\gamma = 1 - \frac{H(p_{C})}{\log |\mathcal{Y}|},
\label{eq:entropy_confidence}
\end{equation}
where $H(\cdot)$ denotes entropy. By construction, $\gamma \in [0,1]$: a concentrated residual distribution yields low entropy and high confidence ($\gamma \to 1$), while a flat residual distribution yields high entropy and low confidence ($\gamma \to 0$).

The final calibrated score is:
\begin{equation}
S(y) = s^{+}(y) + \gamma \, C(y).
\label{eq:final_score}
\end{equation}
This formulation preserves the positive-profile score as the anchored base and adds a reliability-weighted residual enhancement. When the residual is confident, the model strengthens the profile-specific signal. When the residual is uncertain, the entropy coefficient reduces its influence and the method falls back toward positive-profile prompting.

\subsection{Instantiations}

The proposed decoding rule is instantiated differently for multiple-choice and open-ended QA.

\paragraph{Multiple-choice QA.}
The candidate space $\mathcal{Y}$ is the set of answer options. We compute $s^{+}(y)$, $s^{-}(y)$, and $s^{0}(y)$ for each option $y$, construct the residual $C(y)$, and obtain the calibrated score $S(y)$. The final answer is $\hat{y} = \arg\max_{y \in \mathcal{Y}} S(y)$.

\paragraph{Open-ended QA.}
We apply the same principle at the token level. At each decoding step $t$, let $y_{<t}$ denote the generated prefix and $v$ a candidate next token. We first compute the per-token scores $s_t^{~}(v) = \log p_{\theta}(v \mid x, q, E^{~}, y_{<t})$ under each evidence condition, and define the token-level residual $C_t(v)$ analogously to Eq.\eqref{eq:residual}.

To compute the token-level confidence coefficient $\gamma_t$, we restrict the candidate token space to the top-$K$ tokens under the positive-profile distribution, denoted $V_t$. This avoids diluting the entropy signal with the overwhelming noise of the full vocabulary. Concretely:
\begin{equation}
\gamma_t = 1 - \frac{H(p_{C,t})}{\log K},
\label{eq:token_confidence}
\end{equation}
where $p_{C,t}$ is the residual-induced distribution over $V_t$ (Eq.\eqref{eq:residual_distribution} applied at token level) and $K=24$ in all experiments. The calibrated next-token score is $S_t(v) = s_t^+(v) + \gamma_t \, C_t(v)$, and the next token is selected by $\arg\max_{v \in V_t} S_t(v)$.

Note that token-level residuals are inherently noisier than response-level ones, since a single token reflects far weaker personalization signal than a complete response. The entropy coefficient is therefore particularly important in this setting: it suppresses residual corrections at uncertain decoding steps, allowing generation to remain anchored to the positive-profile baseline when the token-level evidence is unreliable.


\section{Experiments}

\subsection{Experimental Setup}

\paragraph{Benchmarks.}
We evaluate on three personalized multimodal benchmarks: MMPB~\cite{kim2026mmpb} for text-profile-driven preference and identity reasoning, and YoLLaVA~\cite{nguyen2024yo} and MyVLM~\cite{alaluf2024myvlm} for image-reference-based personalized visual recognition. 
We report macro accuracy on MMPB and weighted accuracy on YoLLaVA and MyVLM.

\paragraph{Baselines.}
We compare with four decoding baselines: \textbf{No Profile}, which uses the frozen model without personalized evidence ($E^{0}$); \textbf{E$^{+}$ Prompting}, which directly prompts with the positive profile; and two single-reference residual variants, \textbf{E$^{0}$ Contrast} with $C(y)=s^{+}(y)-s^{0}(y)$ and \textbf{E$^{-}$ Contrast} with $C(y)=s^{+}(y)-s^{-}(y)$. 
Prior personalized VLM systems often differ in how evidence is acquired, retrieved, or learned, so they are not direct decoding baselines under our setting; we provide a contextual comparison in Appendix D.

\paragraph{Implementation details.}
We construct $E^{-}$ from the same dataset: for text profiles, we invert preference semantics where possible and otherwise sample profiles from other users; for visual references, we select same-category images of different subjects. 
We use $\alpha=1.25$, $\beta=1.25$, and $T=1.25$ by default, and compute token-level confidence over the top-$K{=}24$ tokens under $E^{+}$ for open-ended decoding. 
Prompt templates, counterfactual construction details, and hyperparameter sensitivity are provided in Appendix B \& C.




\begin{table*}[t]
\centering
\small
\renewcommand{\arraystretch}{1.25}
\setlength{\tabcolsep}{6pt}
\begin{tabular}{llcccc}
\hline
Model & Method & MMPB (Macro) & YoLLaVA & MyVLM & Avg. \\
\hline 
\multirow{5}{*}{Qwen2.5-VL-3B}
& \cellcolor{gray!15} No Profile & \cellcolor{gray!15} 49.82 & \cellcolor{gray!15} 42.05 & \cellcolor{gray!15} 50.00 & \cellcolor{gray!15} 47.29 \\
& $E^+$ Prompting & 68.98 & 86.67 & 95.86 & 83.84 \\
& $E^0$ Contrast & 69.68$_{\textcolor{green!50!black}{\scriptstyle +0.70}}$ & 89.23$_{\textcolor{green!50!black}{\scriptstyle +2.56}}$ & 96.13$_{\textcolor{green!50!black}{\scriptstyle +0.27}}$ & 85.01$_{\textcolor{green!50!black}{\scriptstyle +1.17}}$ \\
& $E^-$ Contrast & \textbf{70.77}$_{\textcolor{green!50!black}{\scriptstyle +1.79}}$ & 95.90$_{\textcolor{green!50!black}{\scriptstyle +9.23}}$ & 98.62$_{\textcolor{green!50!black}{\scriptstyle +2.76}}$ & 88.43$_{\textcolor{green!50!black}{\scriptstyle +4.59}}$ \\
& \textbf{Ours} & 70.76$_{\textcolor{green!50!black}{\scriptstyle +1.78}}$ & \textbf{97.44}$_{\textcolor{green!50!black}{\scriptstyle +10.77}}$ & \textbf{99.38}$_{\textcolor{green!50!black}{\scriptstyle +3.52}}$ & \textbf{89.19}$_{\textcolor{green!50!black}{\scriptstyle +5.35}}$ \\
\hline 
\multirow{5}{*}{Qwen2.5-VL-7B}
& \cellcolor{gray!15} No Profile & \cellcolor{gray!15} 48.80 & \cellcolor{gray!15} 52.31 & \cellcolor{gray!15} 50.69 & \cellcolor{gray!15} 50.60 \\
& $E^+$ Prompting & 72.30 & 91.28 & 96.87 & 86.82 \\
& $E^0$ Contrast & 72.71$_{\textcolor{green!50!black}{\scriptstyle +0.41}}$ & 91.79$_{\textcolor{green!50!black}{\scriptstyle +0.51}}$ & 96.91$_{\textcolor{green!50!black}{\scriptstyle +0.04}}$ & 87.14$_{\textcolor{green!50!black}{\scriptstyle +0.32}}$ \\
& $E^-$ Contrast & \textbf{73.89}$_{\textcolor{green!50!black}{\scriptstyle +1.59}}$ & 97.95$_{\textcolor{green!50!black}{\scriptstyle +6.67}}$ & 98.18$_{\textcolor{green!50!black}{\scriptstyle +1.31}}$ & 90.01$_{\textcolor{green!50!black}{\scriptstyle +3.19}}$ \\
& \textbf{Ours} & 72.91$_{\textcolor{green!50!black}{\scriptstyle +0.61}}$ & \textbf{98.97}$_{\textcolor{green!50!black}{\scriptstyle +7.69}}$ & \textbf{98.83}$_{\textcolor{green!50!black}{\scriptstyle +1.96}}$ & \textbf{90.24}$_{\textcolor{green!50!black}{\scriptstyle +3.42}}$ \\
\hline 
\multirow{5}{*}{Qwen3-VL-2B}
& \cellcolor{gray!15} No Profile & \cellcolor{gray!15} 48.93 & \cellcolor{gray!15} 35.90 & \cellcolor{gray!15} 51.12 & \cellcolor{gray!15} 45.32 \\
& $E^+$ Prompting & 70.99 & 86.15 & 96.25 & 84.46 \\
& $E^0$ Contrast & 71.25$_{\textcolor{green!50!black}{\scriptstyle +0.26}}$ & 89.23$_{\textcolor{green!50!black}{\scriptstyle +3.08}}$ & \textbf{97.12}$_{\textcolor{green!50!black}{\scriptstyle +0.87}}$ & 85.87$_{\textcolor{green!50!black}{\scriptstyle +1.41}}$ \\
& $E^-$ Contrast & 72.72$_{\textcolor{green!50!black}{\scriptstyle +1.73}}$ & 97.95$_{\textcolor{green!50!black}{\scriptstyle +11.80}}$ & 96.36$_{\textcolor{green!50!black}{\scriptstyle +0.11}}$ & 89.01$_{\textcolor{green!50!black}{\scriptstyle +4.55}}$ \\
& \textbf{Ours} & \textbf{73.03}$_{\textcolor{green!50!black}{\scriptstyle +2.04}}$ & \textbf{98.46}$_{\textcolor{green!50!black}{\scriptstyle +12.31}}$ & 97.04$_{\textcolor{green!50!black}{\scriptstyle +0.79}}$ & \textbf{89.51}$_{\textcolor{green!50!black}{\scriptstyle +5.05}}$ \\
\hline 
\multirow{5}{*}{InternVL3-2B}
& \cellcolor{gray!15} No Profile & \cellcolor{gray!15} 49.94 & \cellcolor{gray!15} 41.03 & \cellcolor{gray!15} 52.37 & \cellcolor{gray!15} 47.78 \\
& $E^+$ Prompting & 71.02 & 51.28 & 79.15 & 67.15 \\
& $E^0$ Contrast & 72.15$_{\textcolor{green!50!black}{\scriptstyle +1.13}}$ & 73.33$_{\textcolor{green!50!black}{\scriptstyle +22.05}}$ & 82.44$_{\textcolor{green!50!black}{\scriptstyle +3.29}}$ & 75.97$_{\textcolor{green!50!black}{\scriptstyle +8.82}}$ \\
& $E^-$ Contrast & \textbf{72.47}$_{\textcolor{green!50!black}{\scriptstyle +1.45}}$ & 76.41$_{\textcolor{green!50!black}{\scriptstyle +25.13}}$ & \textbf{94.52}$_{\textcolor{green!50!black}{\scriptstyle +15.37}}$ & 81.13$_{\textcolor{green!50!black}{\scriptstyle +13.98}}$ \\
& \textbf{Ours} & 72.03$_{\textcolor{green!50!black}{\scriptstyle +1.01}}$ & \textbf{87.18}$_{\textcolor{green!50!black}{\scriptstyle +35.90}}$ & 92.57$_{\textcolor{green!50!black}{\scriptstyle +13.42}}$ & \textbf{83.93}$_{\textcolor{green!50!black}{\scriptstyle +16.78}}$ \\
\hline 
\multirow{5}{*}{InternVL3-8B}
& \cellcolor{gray!15} No Profile & \cellcolor{gray!15} 50.50 & \cellcolor{gray!15} 47.18 & \cellcolor{gray!15} 49.86 & \cellcolor{gray!15} 49.18 \\
& $E^+$ Prompting & 69.04 & 94.87 & 95.60 & 86.50 \\
& $E^0$ Contrast & \textbf{72.32}$_{\textcolor{green!50!black}{\scriptstyle +3.28}}$ & 94.36$_{\scriptstyle -0.51}$ & 98.95$_{\textcolor{green!50!black}{\scriptstyle +3.35}}$ & 88.54$_{\textcolor{green!50!black}{\scriptstyle +2.04}}$ \\
& $E^-$ Contrast & 70.91$_{\textcolor{green!50!black}{\scriptstyle +1.87}}$ & \textbf{98.97}$_{\textcolor{green!50!black}{\scriptstyle +4.10}}$ & 97.94$_{\textcolor{green!50!black}{\scriptstyle +2.34}}$ & 89.27$_{\textcolor{green!50!black}{\scriptstyle +2.77}}$ \\
& \textbf{Ours} & 71.28$_{\textcolor{green!50!black}{\scriptstyle +2.24}}$ & 98.46$_{\textcolor{green!50!black}{\scriptstyle +3.59}}$ & \textbf{99.44}$_{\textcolor{green!50!black}{\scriptstyle +3.84}}$ & \textbf{89.73}$_{\textcolor{green!50!black}{\scriptstyle +3.23}}$ \\
\hline 
\end{tabular}
\caption{Overall performance across models and personalized multimodal benchmarks. ``No Profile'' denotes the no-personalization lower bound using the empty-profile condition. Green subscripts indicate absolute gains over $E^+$ Prompting. ``$E^0$ Contrast'' and ``$E^-$ Contrast'' are simple single-reference residual baselines without entropy calibration. We report macro accuracy on MMPB and weighted accuracy on YoLLaVA and MyVLM. Avg.\ denotes the arithmetic mean over the three benchmarks.}
\label{tab:main_results}
\end{table*}

\subsection{Main Results}

Table~\ref{tab:main_results} reports the overall performance across three benchmarks and five backbones.

\paragraph{Positive prompting provides a strong but incomplete baseline.}
Direct $E^{+}$ prompting already yields large gains over the no-profile model, confirming that current VLMs can exploit user profiles when explicitly provided. However, $E^{+}$ scores conflate personalized evidence with generic model priors, and the resulting improvement is uneven: on preference-oriented tasks (Table~\ref{tab:mmpb_fine_grained}), $E^{+}$ prompting remains weak (e.g., only 32.85 on Pref.~Aware), indicating that profile-supported and prior-driven responses are not yet well separated.

\begin{figure}[t]
\centering
\includegraphics[width=0.42\textwidth]{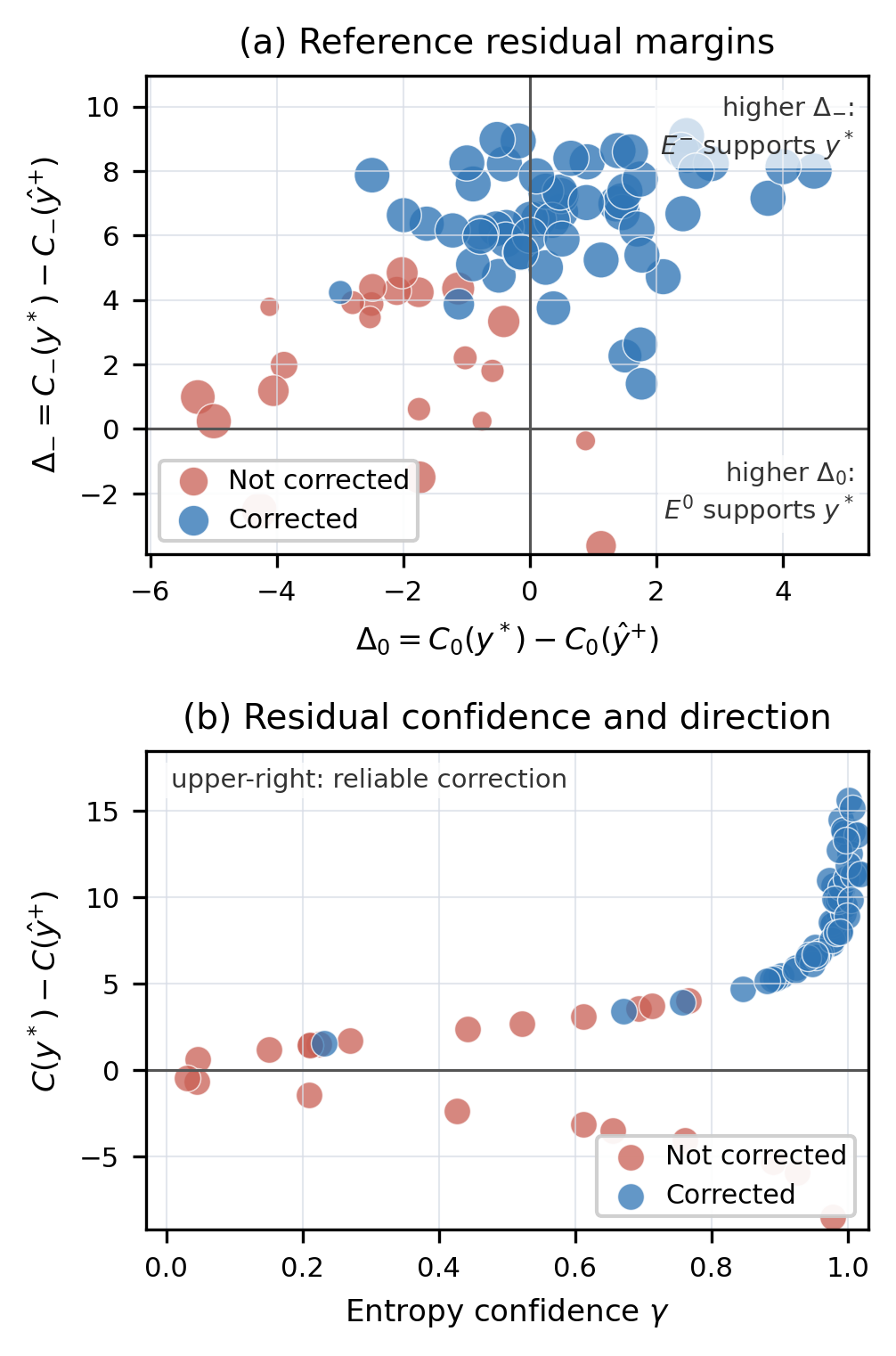}
\caption{
Residual evidence analysis.
Each point corresponds to an error made by direct $E^{+}$ prompting; blue points are corrected by our method, while red points remain incorrect.
Panel (a) compares the residual margins provided by the empty-profile reference $E^{0}$ and the counterfactual reference $E^{-}$, where larger values indicate stronger support for the gold answer over the original $E^{+}$ prediction.
Panel (b) relates entropy confidence $\gamma$ to the residual margin for the gold answer, showing whether the residual is both reliable and directed toward the correct response.
Point size in panel (a) indicates $\gamma$.
}
\label{fig:residual_scatter_analysis}
\end{figure}

\paragraph{Counterfactual contrast is a stronger reference than empty-profile contrast.}
Comparing the two single-reference baselines, $E^{-}$ Contrast consistently and substantially outperforms $E^{0}$ Contrast across nearly all model--benchmark pairs. For instance, on YoLLaVA with Qwen2.5-VL-3B, $E^{-}$ Contrast achieves 95.90 while $E^{0}$ Contrast achieves only 89.23 (+6.67). This confirms that contrasting against a concrete alternative user ($E^{-}$) exposes personalized evidence more effectively than contrasting against the mere absence of evidence ($E^{0}$), supporting our design of jointly using both references.
The two references provide complementary signals. 
While $E^{0}$ exposes the model's generic prior in the absence of personalization, $E^{-}$ tests whether the same response would remain preferred for an alternative user. 
The stronger performance of $E^{-}$ Contrast suggests that counterfactual evidence is particularly useful for isolating user-specific signals, and we further analyze this effect in Figure~\ref{fig:residual_scatter_analysis}.

\paragraph{Entropy-calibrated decoding provides the most reliable performance.}
Our full method achieves competitive performance on identity-sensitive visual tasks: on YoLLaVA, it outperforms $E^{+}$ prompting by +3.59 to +35.90 across models; on MyVLM, it reaches 99\% on two of the five models. On the text-profile benchmark MMPB, our method closely tracks or matches $E^{-}$ Contrast, with the entropy coefficient providing conservative modulation when the residual is diffuse. Fine-grained analysis (Table~\ref{tab:mmpb_fine_grained}) shows the largest gains on preference-oriented sub-tasks: Pref.~Aware improves from 32.85 to 43.00 (+10.15), and Pref.~Incon. from 73.90 to 90.00 (+16.10). On recognition-oriented sub-tasks, gains are more moderate but consistent. Taken together, the results confirm that training-free personalization benefits from two complementary ingredients: anchoring on the positive-profile response, and adaptively calibrating the residual through contrastive evidence and uncertainty-aware modulation.


    \begin{table}[t]
    \centering
    \small
    \renewcommand{\arraystretch}{1.25}
    \setlength{\tabcolsep}{5pt}
    \begin{tabular}{lcccc}
    \hline
    Method & P.A. & P.I. & R.A. & R.I. \\
    \hline
    $E^+$ Prompting & 32.85 & 73.90 & 76.80 & \textbf{81.95} \\
    $E^0$ Contrast & 40.90 & 83.50 & 81.08 & 79.40 \\
    $E^-$ Contrast & 39.65 & 88.00 & \textbf{94.29} & 80.90 \\
    \textbf{Ours} & \textbf{43.00} & \textbf{90.00} & 94.22 & 79.55 \\
    \hline
    \end{tabular}
    \caption{Fine-grained results on representative MMPB sub-tasks. Pref.Aware(P.A.) and Pref.Incon.(P.I.) evaluate preference-oriented personalization, while Rec.Aware(R.A.) and Rec.Incon.(R.I.) evaluate personalized visual recognition under user-specific context.}
    \label{tab:mmpb_fine_grained}
    \end{table}

    \begin{table}[t]
    \centering
    \small
    \renewcommand{\arraystretch}{1.25}
    \setlength{\tabcolsep}{6pt}
    \begin{tabular}{lccc}
    \hline
    Method & Accuracy & $\Delta$ vs Base \\
    \hline
    \rowcolor{gray!12}
    Base & \textbf{84.84} & 0.00 \\
    w/o Entropy & 75.23 & -9.61 \\
    Ours & 81.59 & -3.25 \\
    \hline
    \end{tabular}
    \caption{Robustness on the public MME benchmark under irrelevant textual personalization context. \textit{Base} denotes the original model without any user profile. \textit{Ours w/o Entropy} removes entropy-based confidence calibration, while \textit{Ours} uses the full calibrated residual decoding rule.}
    \label{tab:mme_text_interference}
    \end{table}

    \begin{table}[t]
    \centering
    \small
    \renewcommand{\arraystretch}{1.25}
    \setlength{\tabcolsep}{6pt}
    \begin{tabular}{lcc}
    \hline
    Method & MMPB (Macro) & YoLLaVA \\
    \hline
    $E^+$ Prompting & 68.98 & 86.67 \\
    w/o Entropy & 70.35 & 95.39 \\
    \textbf{Ours} & \textbf{70.76} & \textbf{97.44} \\
    \hline
    \end{tabular}
    \caption{Ablation of entropy-based confidence calibration on personalized benchmarks. \textit{Ours w/o Entropy} removes confidence calibration and directly applies residual personalization, while \textit{Ours} uses the full entropy-calibrated decoding rule.}
    \label{tab:entropy_ablation}
    \end{table}

\subsection{Ablation Study}

We ablate the reference evidence conditions and entropy-based confidence calibration. 
As shown in Table~\ref{tab:main_results}, $E^{-}$ Contrast consistently outperforms $E^{0}$ Contrast across most model--benchmark pairs, suggesting that a concrete counterfactual profile provides a stronger discriminative signal than the absence of personalization, while $E^{0}$ still anchors the residual against generic priors. 
For entropy calibration, we evaluate two complementary settings. 
On the public MME benchmark with an irrelevant textual profile prepended to each sample~\cite{fu2026mme}, uncalibrated residual decoding degrades the base model by $-9.60$ (Table~\ref{tab:mme_text_interference}), showing that noisy profile evidence can be over-amplified. 
The full method reduces this drop to $-3.25$. 
On personalized benchmarks (Table~\ref{tab:entropy_ablation}), entropy calibration further improves or preserves the gains of residual decoding. 
Together, these results indicate that entropy calibration makes the residual reliability-aware rather than uniformly amplified.

\subsection{Component Analysis}

Beyond aggregate ablations, we further analyze how the reference conditions and entropy confidence affect individual predictions.
For each error made by direct $E^{+}$ prompting, we compute reference residual margins $\Delta_0=C_0(y^*)-C_0(\hat{y}^{+})$ and $\Delta_-=C_-(y^*)-C_-(\hat{y}^{+})$, where $C_0=s^{+}-s^{0}$ and $C_-=s^{+}-s^{-}$.
Here, $E^{0}$ measures how much the model favors a response without personalized evidence, while $E^{-}$ tests whether the same response remains preferred under an alternative user profile.
As shown in Figure~\ref{fig:residual_scatter_analysis}(a), corrected errors often have large positive $\Delta_-$, suggesting that the counterfactual reference provides discriminative user-specific evidence beyond the generic prior exposed by $E^{0}$.
The cluster of corrected samples in the upper-right region further indicates that the two references can provide complementary support, yielding a more stable residual signal when they agree on the correction direction.
Figure~\ref{fig:residual_scatter_analysis}(b) further shows that corrected errors are concentrated in the upper-right region, where high entropy confidence coincides with a positive residual margin for the gold answer. This suggests that the entropy coefficient provides a useful proxy for the reliability of residual-based correction.


\subsection{Open-Ended Personalized QA}

    \begin{table}[t]
    \centering
    \small
    \renewcommand{\arraystretch}{1.25}
    \setlength{\tabcolsep}{6pt}
    \begin{tabular}{lccccc}
    \hline
    Subset  & Win & Tie & Avg.\ Score ($\uparrow$) \\
    \hline
    Overall & 63.0 & 18.0 & 3.54 $\rightarrow$ \textbf{4.70} \\
    Text-profile QA & 45.0 & 30.0 & 3.80 $\rightarrow$ \textbf{4.38} \\
    Image-profile QA & 75.0 & 10.0 & 3.37 $\rightarrow$ \textbf{4.92} \\
    \hline
    \end{tabular}
    \caption{LLM-judge evaluation on the 100-example open-ended personalized QA benchmark using Qwen2.5-VL-3B. We compare standard positive prompting against our token-level calibrated residual decoding. Win/Tie denotes the percentage of cases where the judge prefers our method, considers the two responses tied. Avg.\ Score reports the mean overall judge score on a 0--6 scale.}
    \label{tab:openended_token_judge}
    \end{table}

\subsection{Extension to Open-ended Personalized Generation}

Although our main experiments focus on closed-set personalized reasoning, the proposed decoding framework can also be applied to open-ended personalized generation. To examine this extension, we evaluate token-level calibrated residual decoding on an open-ended personalized QA benchmark.

\paragraph{Evaluation setup.}
We construct a 100-example open-ended personalized QA benchmark from public datasets, with 40 text-profile examples from MMPB and 60 image-profile examples from YoLLaVA and MyVLM. 
We compare direct \textit{Positive Prompting} with \textit{Ours (Token-level)}, which applies calibrated residual decoding at each generation step. 
An external LLM judge~\cite{yang2025qwen3} compares response pairs along factuality, personalization, and counterfactual robustness, each scored from 0 to 2. 
We report win/tie rates and average overall scores.

\paragraph{Open-ended results.}
As shown in Table~\ref{tab:openended_token_judge}, our method achieves a 63.0\% overall win rate and improves the average score from 3.54 to 4.70. 
The gain is strongest on image-profile QA (75.0\% win rate, 3.37$\to$4.92), where personalization is grounded in identity-sensitive visual evidence. 
Text-profile QA shows a smaller but positive gain (45.0\% win rate, 3.80$\to$4.38), likely because long textual preference descriptions introduce greater ambiguity in free-form generation.



\begin{figure}[t]
\centering
\includegraphics[width=0.47\textwidth]{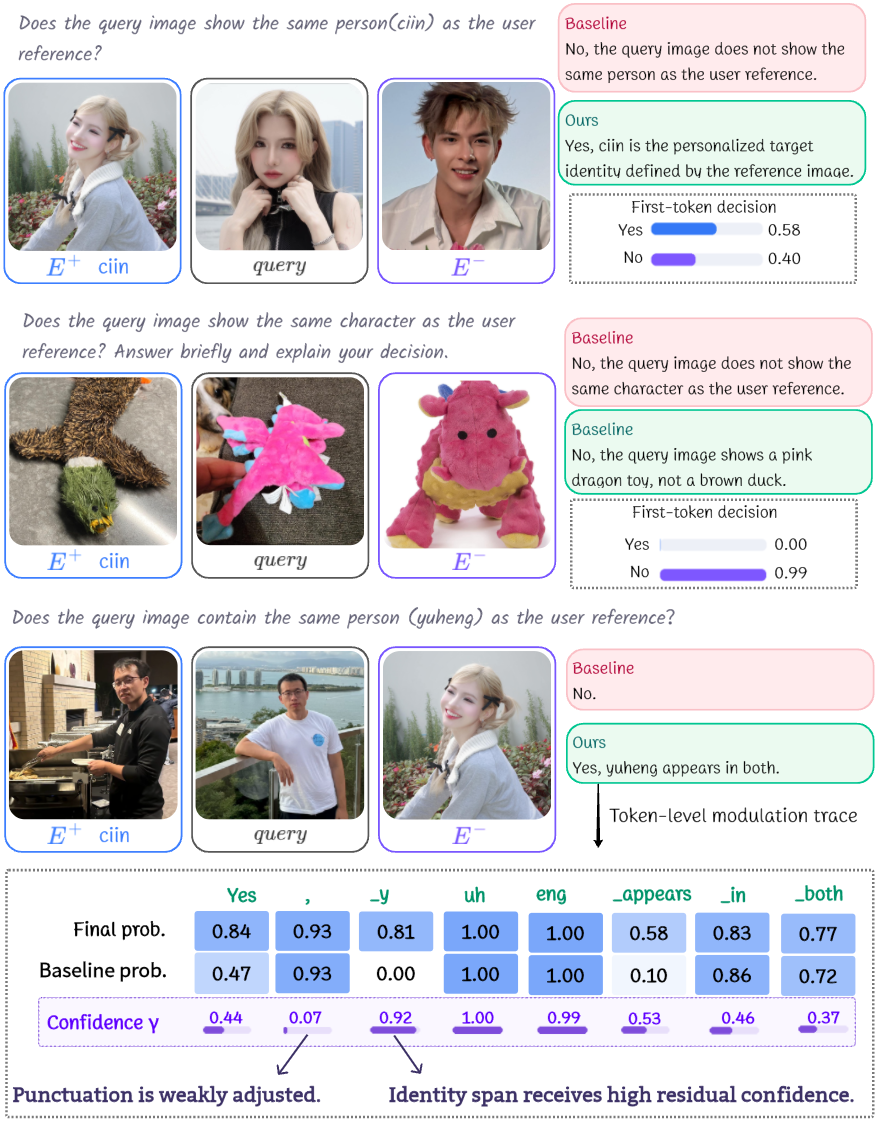}
\caption{
Qualitative examples of token-level personalized decoding.
The calibrated residual shifts the first-token decision toward the profile-supported answer and assigns higher confidence to identity-specific tokens during full-answer generation.
}
\label{fig:openended_cases}
\end{figure}

\paragraph{Qualitative analysis.}
Figure~\ref{fig:openended_cases} illustrates how calibrated residual decoding affects open-ended generation at the token level.
In the first-token decision, our method can shift the model away from the incorrect direct $E^{+}$ prompting response toward the profile-supported answer.
During full-answer generation, the modulation is selective: low-confidence tokens such as punctuation are only weakly adjusted, while identity-specific tokens receive higher residual confidence and larger probability shifts.
This suggests that the proposed decoding rule improves open-ended personalization by strengthening reliable user-specific evidence rather than uniformly amplifying all generated tokens. 
More details on the open-ended QA construction, judge prompt, dimension-level scores, and additional qualitative cases are included in Appendix D \& E.


\section{Conclusion}
We proposed a training-free calibrated residual decoding framework for personalized VLMs, addressing the ambiguity that direct positive-profile prompting can mix genuine user-specific evidence with generic model priors. 
Our method anchors decoding on the positive-profile prediction, estimates the marginal contribution of personalized evidence using counterfactual and empty-profile references, and calibrates the residual strength with entropy-based confidence. 
Experiments across text-profile, image-reference, and open-ended personalized QA settings show consistent gains without model fine-tuning. 
A limitation is the additional inference cost from evaluating multiple evidence conditions, especially in token-level open-ended generation. 
Future work may reduce this overhead through caching, candidate pruning, or selectively applying residual decoding at key decision steps.


\bibliography{aaai2027}


\end{document}